\documentclass[10pt,journal,compsoc]{IEEEtran}

\ifCLASSOPTIONcompsoc
  \usepackage[nocompress]{cite}
\else
  \usepackage{cite}
\fi

\ifCLASSINFOpdf
\else
\fi
\usepackage{hyperref}
\usepackage{amsmath}
\usepackage{graphicx}
\usepackage{ragged2e}
\usepackage{bm}
\usepackage[numbers,sort&compress]{natbib}

\begin{document}

\title{PG\textsuperscript{2}Net: Personalized and Group Preferences Guided Network for Next Place Prediction}

\author{Huifeng Li,~Bin Wang{$^\ast$},~Fan Xia,~Xi Zhai,~Sulei Zhu,~Yanyan Xu{$^\ast$}
\IEEEcompsocitemizethanks{
\IEEEcompsocthanksitem H. Li, B. Wang, S. Zhu are with the College of Information, Mechanical, and Electrical Engineering, Shanghai Normal University, Shanghai 201400, China.E-mail: lhfqy1995@gmail.com, binwang@shnu.edu.cn, suleizhu@163.com.

\IEEEcompsocthanksitem F. Xia is with the School of Information Science and Engineering, Shandong University, Qingdao 266237, China. E-mail: xiafan.1779@foxmail.com.

\IEEEcompsocthanksitem X. Zhai is with the Shanghai Municipal Country-rural Construction and Transportation Institute. E-mail: jessie\_zx28@163.com.

\IEEEcompsocthanksitem Y. Xu is with the MoE Key Laboratory of Artificial Intelligence and AI Institute, Shanghai Jiao Tong University, Shanghai 200240, China. E-mail: yanyanxu@sjtu.edu.cn.

\IEEEcompsocthanksitem This work has been submitted to the IEEE for possible publication. Copyright may be transferred without notice, after which this version may no longer be accessible.

$^\ast${}Corresponding \ authors.
}
}


\IEEEtitleabstractindextext{%

\begin{abstract}
\justifying
Predicting the next place to visit is a key in human mobility behavior modeling, which plays a significant role in various fields, such as epidemic control, urban planning, traffic management, and travel recommendation. To achieve this, one typical solution is designing modules based on RNN to capture their preferences to various locations. Although these RNN-based methods can effectively learn individual's hidden personalized preferences to her visited places, the interactions among users can only be weakly learned through the representations of locations. Targeting this, we propose an end-to-end framework named personalized and group preference guided network (PG\textsuperscript{2}Net), considering the users' preferences to various places at both individual and collective levels. Specifically, PG\textsuperscript{2}Net concatenates Bi-LSTM and attention mechanism to capture each user's long-term mobility tendency. To learn population’s group preferences, we utilize spatial and temporal information of the visitations to construct a spatio-temporal dependency module. We adopt a graph embedding method to map users' trajectory into a hidden space, capturing their sequential relation. In addition, we devise an auxiliary loss to learn the vectorial representation of her next location. Experiment results on two Foursquare check-in datasets and one mobile phone dataset indicate the advantages of our model compared to the state-of-the-art baselines. Source codes are available at \href{https://github.com/urbanmobility/PG2Net}{https://github.com/urbanmobility/PG2Net}.

\end{abstract}

\begin{IEEEkeywords}
Next place prediction, human mobility, trajectory data, personalized and group preferences, attention mechanism.
\end{IEEEkeywords}}

\maketitle

\IEEEdisplaynontitleabstractindextext

\IEEEpeerreviewmaketitle

\IEEEraisesectionheading{\section{Introduction}\label{sec:introduction}}

\IEEEPARstart {W}{ith} the rapid development of information and communication technologies, users can share their locations almost anytime and anywhere to acquire the location-aware services, which forms an abundant amount of trajectory data. Location-based social networks (LBSNs), such as Foursquare and Yelp, collect a huge amount of location data from millions of individuals~\cite{zhang2020next,xu2021understanding}. These trajectory data provide an unprecedented opportunity to study human mobility behavior at scale~\cite{gonzalez2008understanding,jiang2016timegeo}. With massive mobile phone data, Gonz{\'a}lez et al. found a high degree of temporal and spatial regularity in human trajectories~\cite{gonzalez2008understanding}. Their radius of gyrations clearly follow a power law distribution, indicating the sample reproducible patterns of human mobility. In a following study, Song et al. measured the potential predictability of human mobility as 93\%~\cite{song2010limits}. Due to the high predictability of human mobility, researchers attempted to model the mobility behavior of the population at urban scale and utilized it to tackle various urban challenges, including traffic congestion mitigation~\cite{ccolak2016understanding,xu2017collective,olmos2018macroscopic}, air pollution exposure estimation~\cite{xu2019unraveling}, planning of electric vehicles charging behavior~\cite{xu2018planning}, control of the epidemics~\cite{tizzoni2014use,chang2020mobility}.

Prediction of user's next place to visit is a key in modeling of human mobility, and has attracted increasing attention from researchers~\cite{alhasoun2017city}. The task aims to predict the destination of next trip for each user. Assuming that a user's next destination is highly correlated with her recently visited locations, Rendle et al. developed a personalized transition matrix based on the Markov chain to capture the influence of users’ recently visited venues on the current mobility decision~\cite{rendle2010factorizing}. Researchers also notice that the users’ periodical behavior observed from long-term history trajectory plays a critical role in the decision of next travel~\cite{wu2019long,jiang2020predicting}. For instance, in a user's daily routine, she or he is accustomed to going to the library every weekend. Therefore, combining the influence of users’ long-term and short-term trajectory can benefit the prediction of next location~\cite{feng2018deepmove, manotumruksa2017deep}. Feng et al. proposed DeepMove based on an attention mechanism to extract the influence of the user’s history trajectory information on the current situation and provide a personalized recommendation on locations to users~\cite{feng2018deepmove}. Manotumruksa et al. proposed a deep recurrent collaborative filtering framework (DRCF) to utilize the location of users who are similar in the historical trajectory to assist the next location prediction. More recently, researchers proposed the CNN-based method and gated network under the RNN-based framework to extract both long-term and short-term preferences of users~\cite{chen2020predicting,sun2020go}. 

\begin{figure*}[htpb]
    \centering
    \includegraphics[width=14cm]{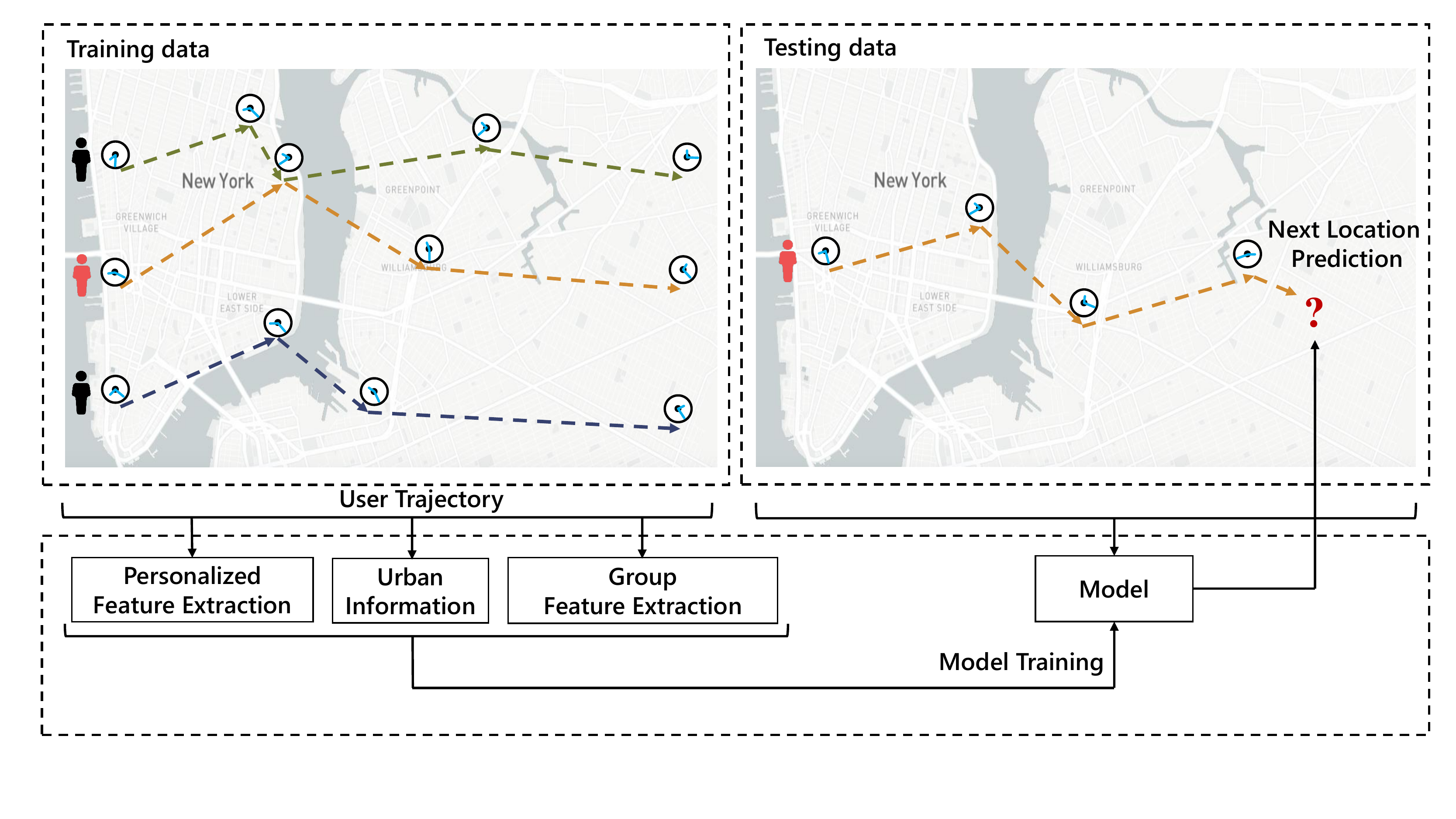}
    \caption{An illustration of next location prediction.}
    \label{fig.1}
\end{figure*}

Although the prediction accuracy has been gradually improved in the fore-mentioned literature, predicting the mobility behavior of large number of users in complex urban environments is still significantly challenging.
On one hand, existing studies focused on modeling the mobility patterns at individual level. The collective pattern of the interaction between the population and the space is not clearly modeled in these methods. In the early studies, statistical physicists used historical movement data to discover the collective pattern of humans. For example, Schneider et al.~\cite{schneider2013unravelling} found that more than 90\% of residents’ mobile behavior conformed to one of 17 basic network modes (Motif). Alessandretti et al.~\cite{alessandretti2018evidence} found that the number of places visited repeatedly by individuals is stable, revealing people’s stable social relations. These studies suggest that statistical physical characteristics are valuable for studying the movement trend of human group level.

On the other hand, existing methods for next location prediction usually cascade the embedding of user ID with the latent vector of locations in the long-term historical and recent trajectories to capture the user's personalized preferences~\cite{guo2020attentional,lian2020geography,liang2020learning}. While users have various preferences to different locations, these methods can not directly model such heterogeneity and dynamic change of the preferences. Targeting this weakness, researchers introduced attention mechanism to learn the personalized preference of each user~\cite{wu2020personalized}. However, users' mobility pattern in space and time is not fully explored from their trajectory sequences. In specific, they focus mainly on exploiting the sequential patterns of user personalised preferences without considering the associated timestamps and geographic information that reflect the collective mobility patterns. Therefore, how to reasonably use trajectory data to learn user's personalized preferences is also the main content of our research.

\begin{figure*}[htpb]
    \centering
    \includegraphics[width=15cm]{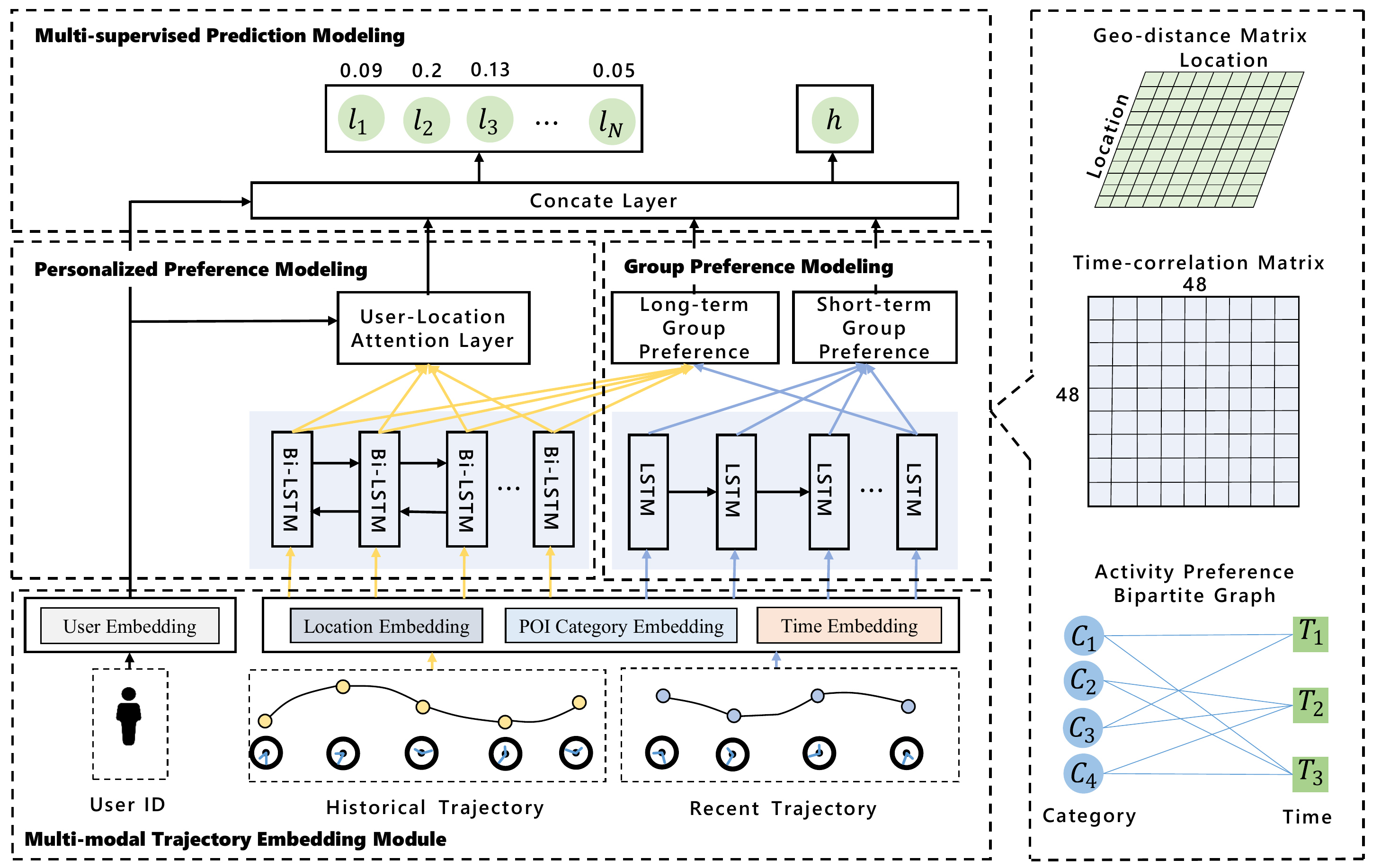}
    \caption{The overall framework of our model.}
    \label{fig.2}
\end{figure*}

To this end, we propose a novel personalized and group preference guided network (PG\textsuperscript{2}Net) to tackle the above issues. The framework is devised to predict each user's next place to visit via considering her preferences to various locations at both individual and collective level. Fig.~\ref{fig.1} illustrates the end-to-end procedure of our learning-based model. PG\textsuperscript{2}Net first learns users' personalized features, as well as the group characteristics from the training data, and integrates the urban information into the prediction module. In the testing phase, the well-trained model is used to predict the next location given their current trajectories. 
Specifically, we consider several key factors that impact the collective mobility behavior, including geographic environment, distance attenuation, and individual spatial activity characteristics. Among them, the geographic environment determines the spatial distribution of potential visit locations which users can reach; the distance attenuation indicates the relationship between user's visit frequency and distance; the individual spatial activity characteristics reflect people’s potential life habits. Considering the above factors, we design a module that combines prior statistical information and recurrent neural network, namely, the dynamic spatio-temporal dependency module, which can learn the influence of group preferences on users' mobile patterns. For learning personalized preferences of each user, we propose a module based on bidirectional long short-term memory networks (Bi-LSTM)~\cite{huang2015bidirectional} and an attention mechanism to capture her dynamic preference. Besides, we propose a novel graph embedding method to represent each location and its categories, which can efficiently learn the sequential relation between visited places. It should be mentioned that most previous studies did not take category information into account. In our model, we attempt to use category information to construct a statistical model to study the characteristics of user group activities. Finally, we propose a novel auxiliary loss function to learn the vectorial representation of the target location and improve the prediction accuracy. The network structure is shown in Fig.~\ref{fig.2}.

Our contributions are summarized as follows:
\begin{enumerate}
\item[(1)] We propose a PG\textsuperscript{2}Net framework to learn the user's personalized preferences and group preferences and predict the next place to visit. PG\textsuperscript{2}Net framework consists of two parts: (i) dynamic spatio-temporal dependency module that leverages temporal and spatial information to model the group preferences of users; (ii) personalized preference module that uses Bi-LSTM and attention mechanism to capture users’ personalized preferences.

\item[(2)] For seeking more efficient hypothesis space for location embedding, we devise an auxiliary loss to enhance the similarity between the representations of the predicted and actual next locations. Consequently, the final loss function is designed to increase the probability of the actual location and decrease the distance between the predicted and actual locations in the hypothesis space.

\item[(3)] We design a group preference module in PG\textsuperscript{2}Net to model the people's collective preferences to variant locations during long-term and short-term periods. PG\textsuperscript{2}Net captures the long-term and short-term preferences via integrating the prior statistical knowledge of people's mobility behavior with the encoding of historical and recent trajectories, respectively.

\item[(4)] We conduct extensive experiments on three real-world datasets in different countries, including two public check-in datasets and a mobile phone dataset, a.k.a. call detail records (CDRs). Experiments show that our model achieves significant improvements over the state-of-the-art methods.
\end{enumerate}


\section{Related Work}

The purpose of the next location prediction task is to recommend a set of ranked locations for users, where the highest-ranking is the predicted value of the next location. At present, there are two types of methods for this task -- conventional (non-deep learning) machine learning-based methods and deep learning-based methods~\cite{feng2018deepmove}.

\subsection{Conventional machine learning-based methods}

The trajectory prediction task needs to mine the users' past trajectory information to predict where the users will go. Typically, for this sequence prediction task, we can use a method based on the Markov model and its variants to make predictions. The Markov-based method is mainly to calculate the transition matrix of the location. According to transition matrix, we can predict where the user will go next. For example, Rendle et al.~\cite{rendle2010factorizing} proposed the factorized personalized Markov chains (FPMC), which is a method applying personalized Markov and matrix factorization to learn users’ transition matrix and overall preference. Cheng et al.~\cite{cheng2013you} proposed a method FPMC-LR to obtain users’ personalized preference and realize the prediction of the next location. Excepted for Markov-based model, Alhasoun et al. utilized information from similar strangers for next place Prediction~\cite{alhasoun2017city}. In this model, they proposed several human mobility similarity metrics used to identify other users with similar mobility characteristics, and proposed dynamic Bayesian network (DBN) model that incorporates the mobility patterns of similar strangers towards better predicting next locations.

\subsection{Deep learning-based methods}
In recent years, deep learning has developed rapidly. Especially, RNN-based methods have attracted increasing attention and been successfully applied to many sequential problems, such as natural language processing, voice recognition, image annotation, and machine translation. So far, researchers have tried to use the RNN-based method for travel information prediction and achieve inspiring results~\cite{liu2016predicting,feng2018deepmove,lan2019travel,sun2020go,zhao2020go}. For example, Liu et al.~\cite{liu2016predicting} proposed the ST-RNN network, which takes users’ adjacent time and space information as the input of the RNN module to capture the spatio-temporal influence. Yao et al.~\cite{yao2017serm} established a recursive model of semantic perception (SERM), which learned the embedding of multiple factors (user, location, time) and captured the time and spatial transition regularity of semantic perception.

Recently, the attention mechanism has been widely used in various fields. Feng et al.~\cite{feng2018deepmove} proposed a model named DeepMove based on attention mechanism and recurrent neural network to capture human mobility. In DeepMove, a multi-module embedding method is adopted to convert sparse features (user, location, time) into a dense representation, and then the historical attention module is used to obtain the most relevant historical trajectory information. However, this method fails to capture the user's dynamic personalized preferences and barely considers the temporal and spatial dependence of the actual users. Gao et al.~\cite{gao2019predicting} proposed a variation-attention-based next location prediction model to overcome the sparsity problem of trajectory data. Wu et al.~\cite{wu2020personalized} proposed a model named PLSPL to learn the specific preference for each user. This method attempts to model category information to predict the next location, but the method does not consider the specific regularities of human mobility and the sequence interaction influences. At present, the latest model for predicting the next location is the LSTPM model proposed by~\cite{sun2020go}, which uses context-aware non-local network and geo-dilated RNN to obtain users' long-term and short-term preferences respectively.

\begin{figure*}[htpb]
    \centering
    \includegraphics[width=14cm]{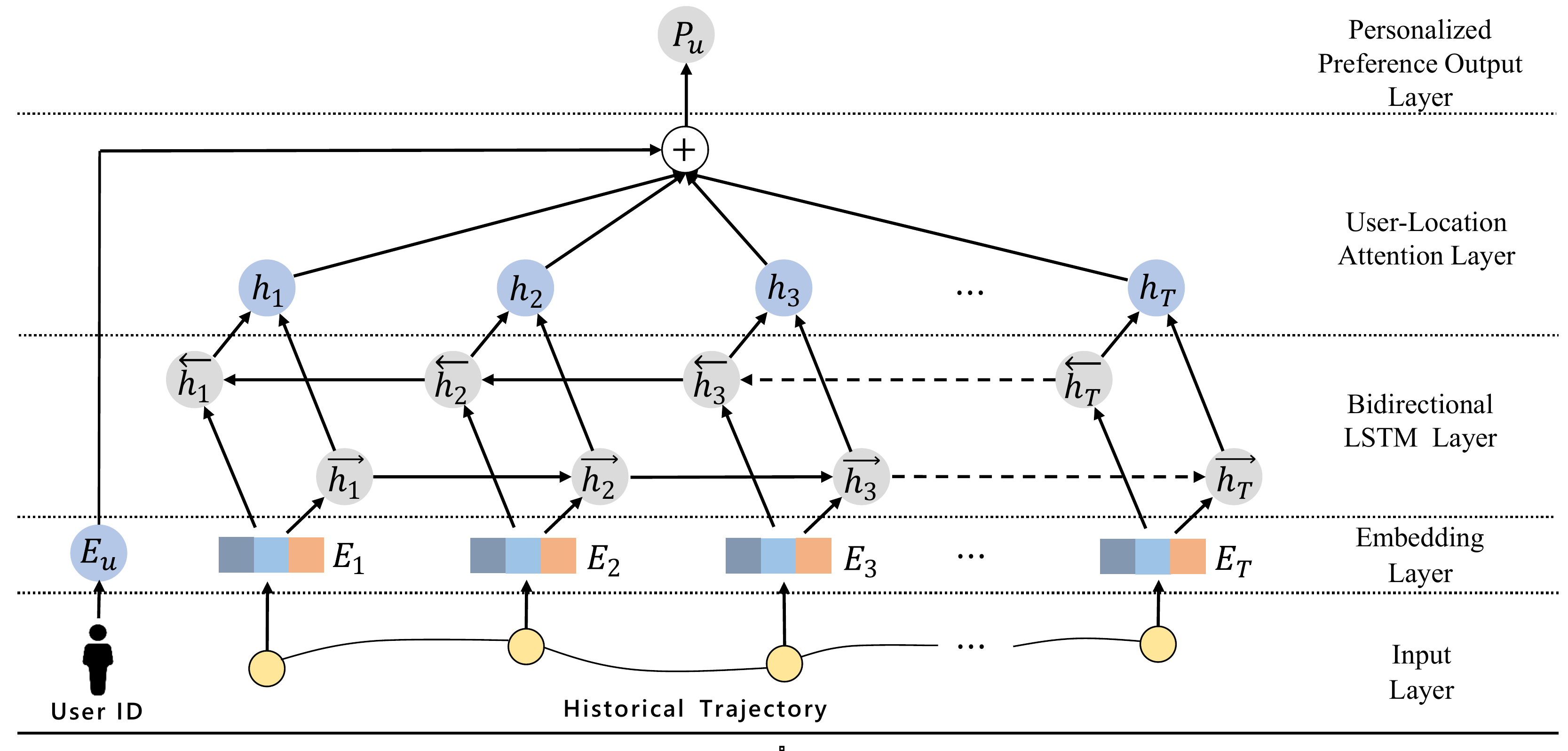}
    \caption{Personalized Preference Modeling.}
    \label{fig.3}
\end{figure*}

\section{Problem Formulation}
\justifying  
We define \textbf{userset}=$\{U_1,U_2,U_3,...,U_n\}$ as a set of LBSN users, and \textbf{locationset}=$\{l_1,l_2,l_3,...,l_m\}$ as a set of trajectories, each of which is geocoded by a (longitude, latitude) tuple representing urban information. Each user contains a large number of trajectories. Taking user $U_1$ as an example, $\{l_2^{U_1},l_5^{U_1},l_4^{U_1},...,l_k^{U_1}\}$ represents all trajectory data of $U_1$ in a period. Because each trajectory corresponds to a category $c_i$ and a timestamp $t_i$, it can be expressed as $\{q_1,q_2,q_3,...,q_k\}$, where $q_i$ contains three attributes $(l_i,c_i,t_i)$. Then we divide each user's trajectories into multiple sub-trajectories in order, for example, $\{S_1^{U_1},S_2^{U_1},S_3^{U_1},...,S_j^{U_1}\}$ is the trajectory sequence of ${U_1}$, where $\{S_1^{U_1},S_2^{U_1},S_3^{U_1},...,S_{j-1}^{U_1}\}$ is his historical trajectory sequence and $S_j$ is his recent trajectory. $S_j$ can be expressed as $\{q_1,q_2,q_3,...,q_{t-1}\}$, where $q_{t-1}$ has three attributes $(l_{t-1},c_{t-1},t_{t-1})$, and represents recent situation. $l_{t-1}$ is the location that $U_1$ has been recently visited. The purpose of this paper is to study the influence of historical trajectories and recent trajectories on the current situation $q_{t-1}$, and predict the top-N of the possible locations of $U_1$ at the next timestamp $t$.
\section{Our model}
In this part, we propose our PG\textsuperscript{2}Net model. We begin with an overview, before zooming into the details.

\subsection{Overview}
The overall framework of PG\textsuperscript{2}Net is depicted in Fig.~\ref{fig.2}. Our PG\textsuperscript{2}Net characterizes the user’s preference to various places at both personalized and group level, and fuses them into a unified framework. Specially, we learn the personalized preference of user \textit{U} from his historical trajectory $\{S_1^{U},S_2^{U},S_3^{U},...,S_{j-1}^{U}\}$, which contains longer trajectory data reflecting the general preference of \textit{U}. And different users have different preferences for the same location. Thus we use user-location attention to learn the latent vectors of user \textit{U} and location $l_i$. Firstly, we learn the latent vectors for user \textit{U} and  POI $q_i$ (which contains location $l_i$, category $c_i$, and timestamp $t_i$) in the multi-modal trajectory embedding module. Then we use Bi-LSTM to learn the historical trajectory’s transition relationship, and we compute the important $a_i$ of each POI $q_i$ to users. Finally, we integrate the sequence information of POIs to present the user’s personalized preference.

As human's decision of moving is impacted by her periodic lifestyle and recent travel behavior, we use historical trajectory to capture her lifestyle while recent trajectory to capture instant decision. Specifically, we firstly learn the latent vectors for $S_j^{U}=\{q_1,q_2,q_3,...,q_{t-1}\}$ in the embedding layer, where $q_i$ contains the location $l_i$, category $c_i$, and timestamp $t_i$. To better understand user’s check-in behaviors, we feed the concatenated embeddings of $(l_i,c_i,t_i)$ into Long Short-Term Memory (LSTM). Then we use a statistical spatio-temporal physical module to model a user's historical trajectory and recent trajectory to learn his group preferences. Finally, concat layer is used to combine the outputs of the personalized preference, long and short-term group preference and feed them into the output layer to generate the final probabilities of candidate locations. It is a remarkable fact that we proposed a novel graph embedding method to learn the latent vectors of locations and categories. At the same time, in the output layer, we propose an auxiliary loss function to supervise the vectorial representation of the next location and improve the prediction accuracy.

\subsection{Multi-modal Trajectory Embedding Module}
Trajectory sequence usually contains a large amount of human mobility information. Due to the mobile device or the user itself, the trajectory sequence has strong sparseness. Targeting this weakness, we use sequence embedding for this kind of data. For example, in check-in sequence, it contains four different types of attributes, namely user ID, timestamp, location, and location category. We will adopt different embedding methods for these different types of attributes in trajectory sequences.

\textbf{user ID and timestamp.} The original user ID and timestamp cannot be directly inputted into the model. We refer to the embedding method mentioned in~\cite{feng2018deepmove,chen2020next} for these two attributes. As each timestamp $t_i$ is continuous, which is difficult to embed, we map it into discrete hours. Firstly, we divide one week into 48 slots, where 0-23 slots represent weekdays, and 24-47 represent weekends. Then each hour is represented as a one-hot 48-dimensional vector, where the non-zero entry denotes the index for the hour. Because one-hot encoding can’t reflect the correlation between sequences, we transform them into $D_t$ dimensional dense vectors and represent them as $\bm{{\rm V}}^t\in{\bm{{\rm R}}^{48\times{D_t}}}$. For a user ID sequence, we utilize the same embedding method to map it into a dense vector, the dimension of which is $D_u$. The embedding vector is represented as $\bm{{\rm V}}^u\in{\bm{{\rm R}}^{N\times{D_u}}}$.


\textbf{location and location category.} In recent years, graph embedding (also known as network embedding) has been applied to many graph related research areas, such as text classification~\cite{lu2020vgcn}, detecting an anomaly in financial networks or social networks~\cite{khazane2019deeptrax,ding2019deep}, etc. The task of our paper is to predict the user's next location, and all potential locations that the user could reach can construct a graph. Therefore, we attempt to adopt a graph embedding method to learn the location representation. Firstly, we use the training dataset to construct a directed weighted graph, where the direction is sequential, and weight refers to the frequency of the two consecutively visited locations. Then we use the graph embedding method node2vec~\cite{grover2016node2vec} to map each location into a low dimensional vector, the dimension of which is $D_l$. The embedding vector is represented as $\bm{{\rm V}}^l\in{\bm{{\rm R}}^{M\times{D_l}}}$. We adopt the same embedding method for the location category sequence. The embedding vector is represented as $\bm{{\rm V}}^c\in{\bm{{\rm R}}^{K\times{D_c}}}$, the dimension of which is $D_c$ . Through this method, we can capture the characteristics of group mobility patterns and location interaction. In the following network training, location and location category embedding will no longer be trained.

The embedding of each POI (which contains location, location category, and the timestamp) can be represented as:
\begin{equation}
    \bm{{\rm E}}_i = [\bm{{\rm V}}_{i}^l\oplus{\bm{{\rm V}}_{i}^t}\oplus{\bm{{\rm V}}_{i}^c}]
\end{equation}
where $\oplus{}$ denotes concatenation, $\bm{{\rm E}}_i$ represents the latent vector for each POI. Different from~\cite{feng2018deepmove,sun2020go} which only learn the latent vector of location, we further consider the context information such as the category of location and the check-in time.

\subsection{Personalized Preference Modeling}
When modeling personalized preferences, an intuitive idea is to learn each user’s location preference. Motivated by this, we propose a user-location attention structure followed by Bi-LSTM to learn the latent representation $\bm{{\rm P}}_u$ of the target user’s personalized preference. As shown in Fig.~\ref{fig.3}, we firstly embed all POIs in each historical trajectory $S_h\in\{S_1,S_2,S_3,...,S_{j-1}\}$, and user embedding into a low-dimensional vector for a user \textit{U}. Then a Bi-LSTM layer is used to learn each POI’s high-level representation and sequential dependency. Finally, we compute the importance $\bm{{\rm a}}_i$ of each POI $q_i$ to users and integrate the sequence information of POIs to present the user’s personalized preference.

To capture the user’s high-level representations and sequential dependencies of different locations, it is beneficial to learn future trajectories as well as past context. Different from~\cite{feng2018deepmove}, which used RNN-based variant network LSTM process sequences in sequence order and ignore future context, the Bi-LSTM network can exploit information both from the past and the future. This is important to learn user’s personalized preferences.

The calculation process of user's personalized preferences can be summarized as follows:

\begin{equation}
    \mathop{\bm{{\rm h}}_i}\limits ^{\rightarrow} = LSTM(\bm{{\rm E}}_i,{\mathop{\bm{{\rm h}}}\limits ^{\rightarrow}}_{i-1}),i\in{S_h}
\end{equation}

\begin{equation}
    \mathop{\bm{{\rm h}}_i}\limits ^{\leftarrow} = LSTM(\bm{{\rm E}}_i,{\mathop{\bm{{\rm h}}}\limits ^{\leftarrow}}_{i-1}),i\in{S_h}
\end{equation}

\begin{equation}
    \bm{{\rm h}}_i= [\mathop{\bm{{\rm h}}_i}\limits ^{\rightarrow} \oplus{\mathop{\bm{{\rm h}}_i}\limits ^{\leftarrow}} ]
\end{equation}

\begin{equation}
    \bm{{\rm a}}_i = \frac{exp(\bm{{\rm h}}_{i}^T \bm{{\rm V}}^u)}{\Sigma^{k}_{i=1}{exp(\bm{{\rm h}}_{i}^T \bm{{\rm V}}^u)}}
\end{equation}

\begin{equation}
    \bm{{\rm P}}_i = \Sigma{\bm{{\rm a}}_i \bm{{\rm h}}_i}
\end{equation}
where $\bm{{\rm V}}^u$ represents the user's latent vector, $\oplus{}$ denotes concatenation representing the combination of the forward and backward outputs, $\bm{{\rm h}}_i$ represents the hidden information of the user's historical trajectory, $\bm{{\rm a}}_i$ denotes the importance of each POI, $\bm{{\rm P}}_u$ is the final representation of personalized preferences of a user \textit{U}.

\begin{figure*}[htpb]
    \centering
    \includegraphics[width=14cm]{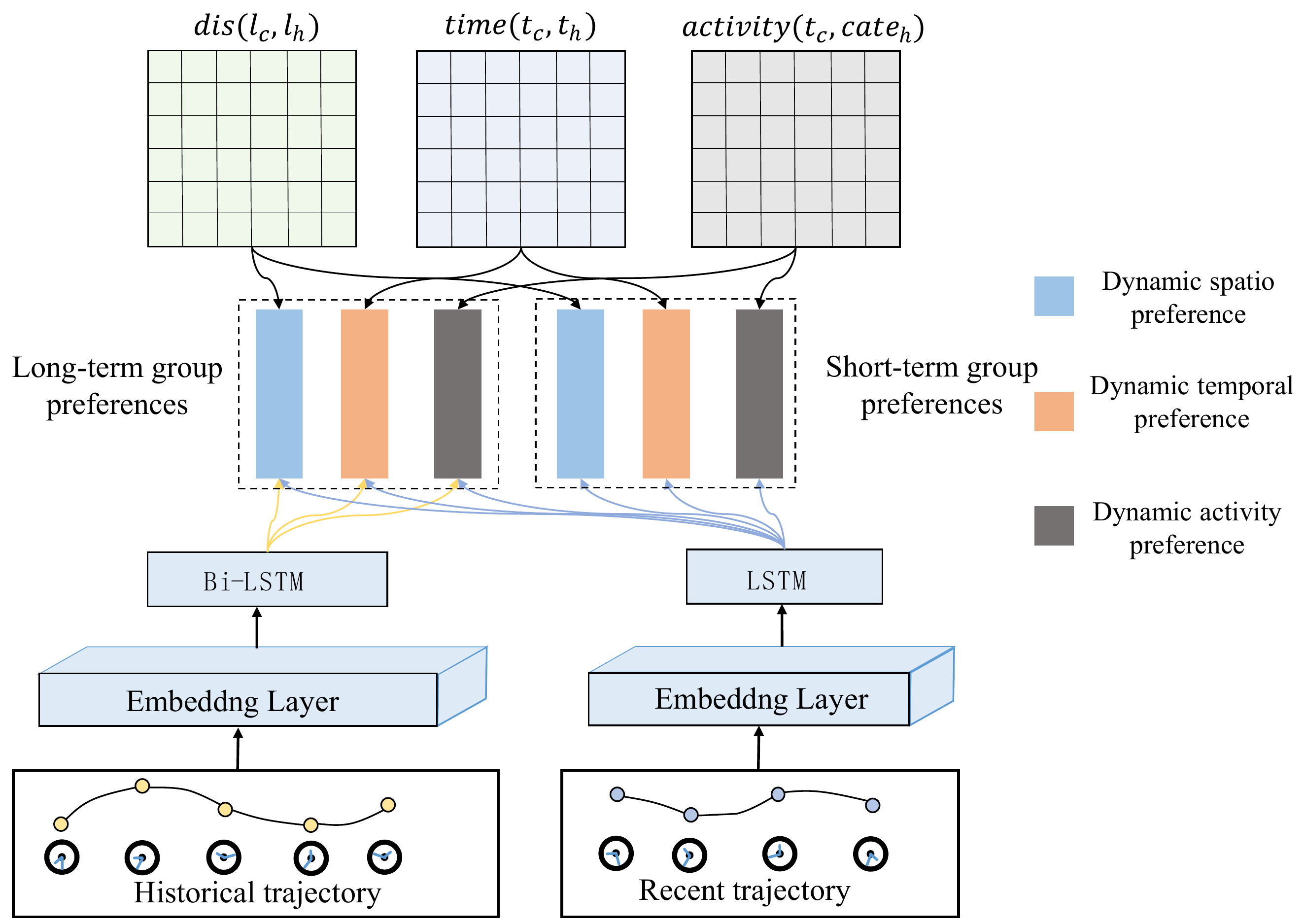}
    \caption{Dynamic Spatio-Temporal Dependency Modeling.}
    \label{fig.4}
\end{figure*}

\subsection{Group Preference Modeling}

When predicting the user's next location, her personalized preferences can reflect the user's general preference. However, the user's interest in the next trajectory not only follows her personalized preference but also is affected by the group behavior pattern. At the same time, the user's preference for the next location changes dynamically with time and space. Therefore, we use trajectory data to construct a statistical physical model named dynamic spatio-temporal dependency module, which uses time and spatial information to model user's group preferences and realizes the prediction of the next location. See Fig.~\ref{fig.4} for the illustration of the proposed model. The dynamic spatio-temporal dependency module is comprised of three parts: the dynamic spatial dependency module, the dynamic time dependency module, and the dynamic activity preference module.

\begin{figure}[htpb]
    \centering
    \includegraphics[width=8cm]{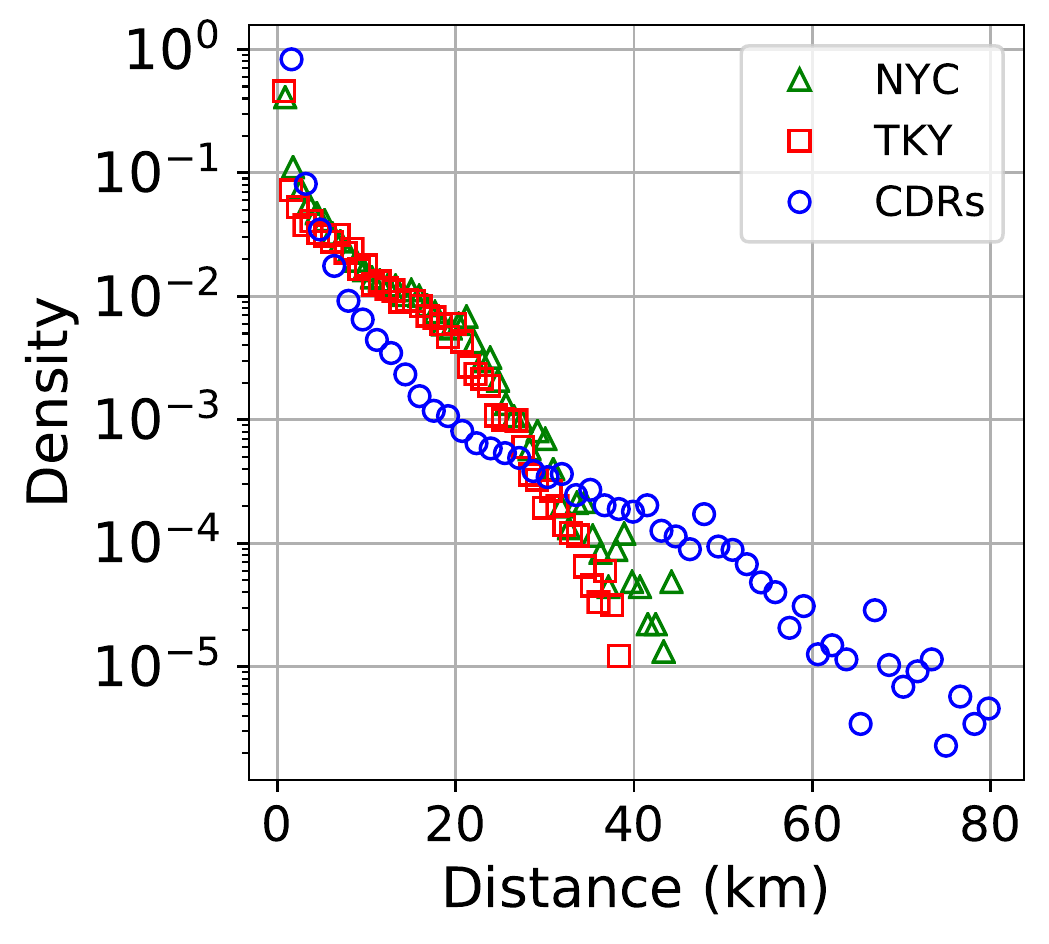}
    \caption{The statistical analysis of the distance factors on users’ spatial preference. Represents the distance distribution between the adjacent location in the three datasets.}
    \label{fig.5}
\end{figure}

\textbf{Dynamic spatial dependency module.} Generally, the distance between geographic locations has a great impact on the user's next location prediction~\cite{sun2020go}. The statistical analysis of the distance between adjacent locations is shown in Fig.~\ref{fig.5}. We can observe that the user's travel patterns completely follow the distance attenuation rule of group mobility. And the higher the cost of long-distance travel, the lower the probability people choose it. That is to say, users tend to visit nearby locations. Based on this observation, this paper proposes a dynamic spatial dependency module to characterize the changing spatial preferences of users when they move. The module can understand the user's dynamic interest in geographic locations, rather than her preference is fixed over time~\cite{zhang2019multi}. When we model the distance preference of users, the key issue is to select the trajectory that has the greatest impact on the recent situation from the history trajectories based on the distance among different locations. Specifically, we firstly generate a geo-distance matrix based on the real-world geographic locations and the historical trajectory data, whose values represent the distance between any locations. Then we generate the weight vector between the recent situation and the historical trajectory based on the distance matrix as follows,

\begin{equation}
    \alpha(l_c,l_h) = \frac{exp^{1/d(l_c,l_k)}}{\Sigma^{n}_{k=1}exp^{1/d(l_c,l_k)}}
\end{equation}
where $d(l_c,l_k)$ is the distance between $l_c$ and $l_k$. Finally, we utilize the above generated weight vector to integrate the sequence information of POIs.

Specifically, for the recent situation $q_{i-1}$ of a user, we first learn its latent embedding vector before modeling spatial preference. Considering that the historical and recent trajectory have different influences on the current situation of users, we utilize geo-distance to model long and short-term spatial group preference,

\begin{equation}
    \bm{{\rm P}}^{s}_{L} = \bm{{\rm H}}_{h}\alpha(l_c,l_h)^{T}
\end{equation}
\begin{equation}
    \bm{{\rm P}}^{s}_{S} = \bm{{\rm H}}_{c}\alpha(l_c,l_h)^{T}
\end{equation}
where $\bm{{\rm H}}_h$ and $\bm{{\rm H}}_c$ are the outputs of Bi-LSTM and LSTM respectively, representing historical trajectory and recent trajectory information. $\bm{{\rm P}}^{s}_L$ and $\bm{{\rm P}}^{s}_S$ represent long and short-term spatial group preference of a user.

\begin{figure}[htpb]
    \centering
    \includegraphics[width=8cm]{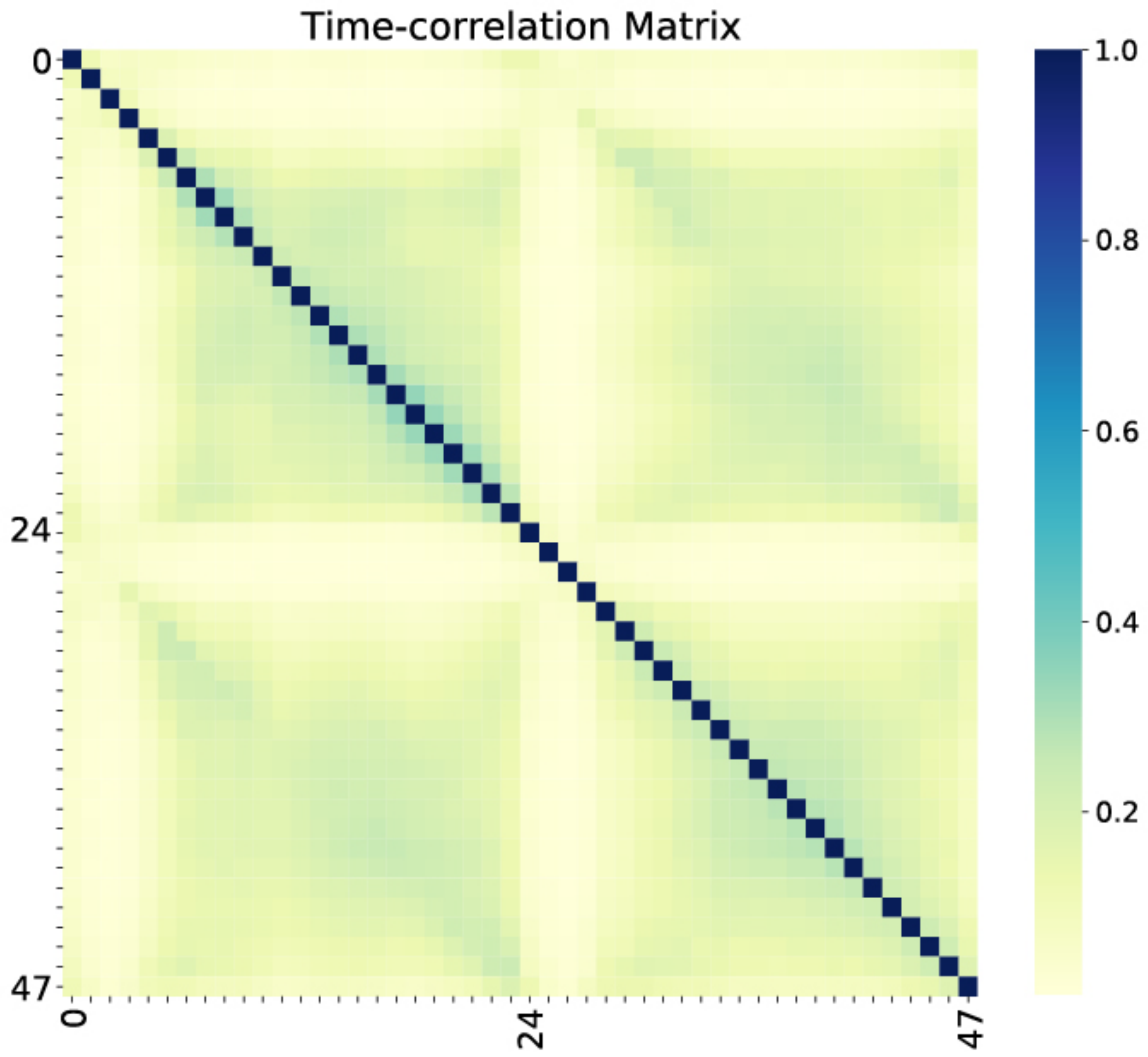}
    \caption{The visualization of the time-correlation matrix.}
    \label{fig.6}
\end{figure}

\textbf{Dynamic time dependency module.} Traditional methods always consider the influence of time factors on the next location prediction~\cite{feng2018deepmove,wu2020personalized}. However, these methods simply learn the semantic relationship of the timestamp sequences and ignore the interaction between the time sequences. For example, most users are accustomed to eating in the cafeteria at 12:00 and 18:00, and drinking coffee in a coffee shop at 15:00 and 20:00. For the users, the location at 12:00 is more related to that at 18:00 rather than 15:00, because both 12:00 and 18:00 are the user’s mealtime. This reflects some group regularities in human movement. Moreover, as the user moves, the user's location preferences at different timestamps are also dynamically changing. Therefore, we propose a dynamic time dependency module to capture the influence of the user's historical trajectory information on the recent state in the time dimension.

We first divide one week into 48 slots, where 0-23 slots represent weekdays, and 24-47 slots represent weekends. We construct a location set to represent the location preference of each slot. For example, $T_i = (l_1,l_3,l_8,...,l_N),i\in{(0,47)}$ means all locations where the i-th slot appears. Then we calculate the time-correlation matrix. As shown in  Fig.~\ref{fig.6}, time correlation of any two slots is expressed as follows,
\begin{equation}
    \Gamma_{i,j} = \frac{\lvert{T_i\cap{T_j}}\rvert}{\lvert{T_i\cup{T_j}}\rvert{}}
\end{equation}
Finally, we generate the weight vector between the recent state and the historical trajectory based on the time-correlation matrix, and utilize the weight vector to integrate the sequence information of POIs,

\begin{equation}
    \alpha{(t_c,t_h)} = \frac{exp^{\Gamma_{c,k}}}{\Sigma^{47}_{k=0}exp^{\Gamma_{c,k}}}
\end{equation}
where $\Gamma_{c,k}$ is the time correlation between c-th and k-th time slots, $\alpha({t_c,t_h})$ is the weight vector between recent state and the historical trajectory. Similar to modeling users' spatial preferences, we utilize time information to model the long and short-term time group preference of the user,

\begin{equation}
    \bm{{\rm P}}^{t}_{L} = \bm{{\rm H}}_{h}\alpha(t_c,t_h)^{T}
\end{equation}

\begin{equation}
    \bm{{\rm P}}^{t}_{S} = \bm{{\rm H}}_{c}\alpha(t_c,t_h)^{T}
\end{equation}
where $\bm{{\rm P}}^{t}_L$ and $\bm{{\rm P}}^{t}_S$ represent long-term and short-term time group preference of the user.

\begin{figure}[htpb]
    \centering
    \includegraphics[width=8cm]{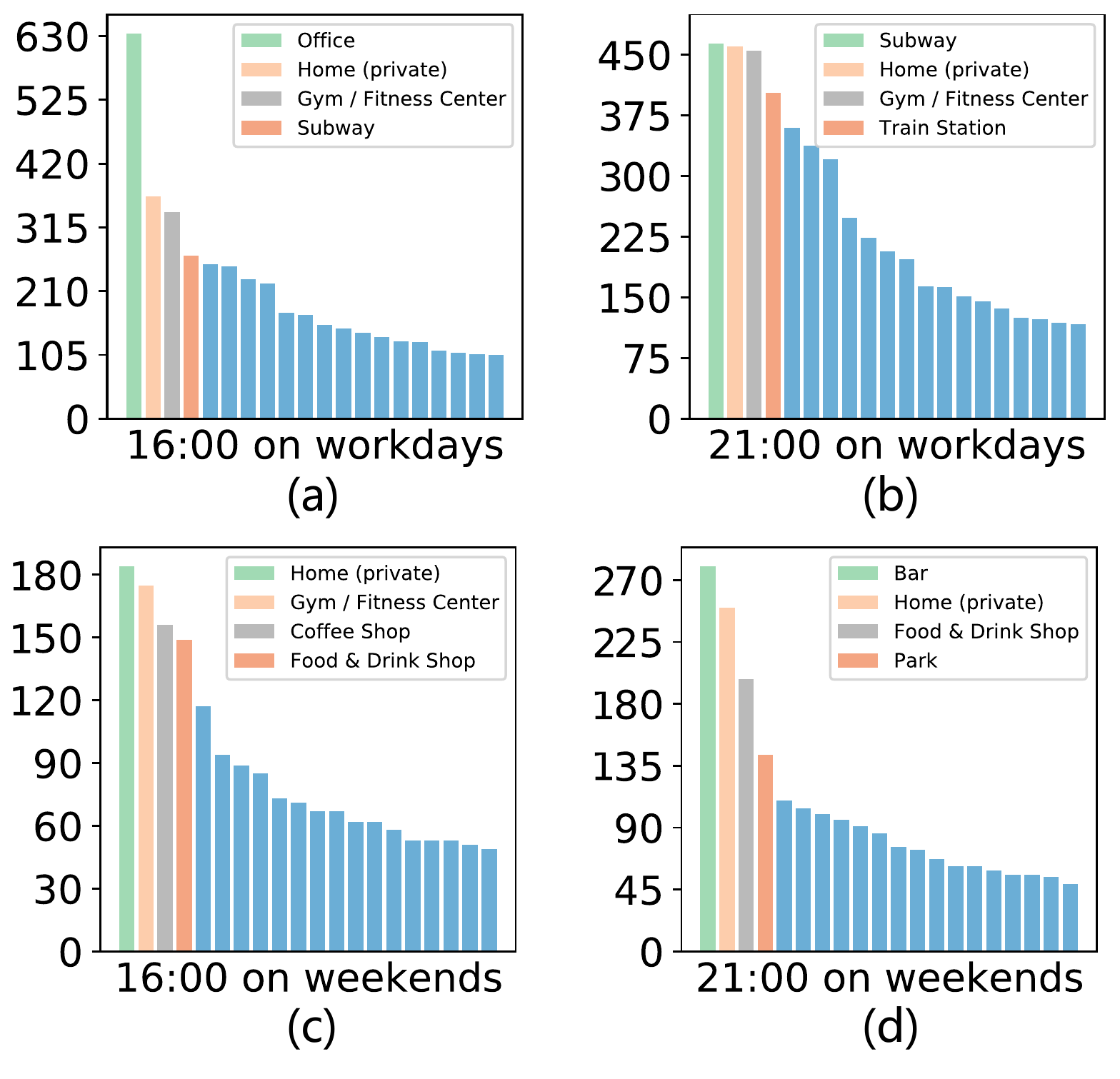}
    \caption{The statistical analysis of location categories in the NYC dataset.}
    \label{fig.7}
\end{figure}

\begin{figure}[htpb]
    \centering
    \includegraphics[width=6cm]{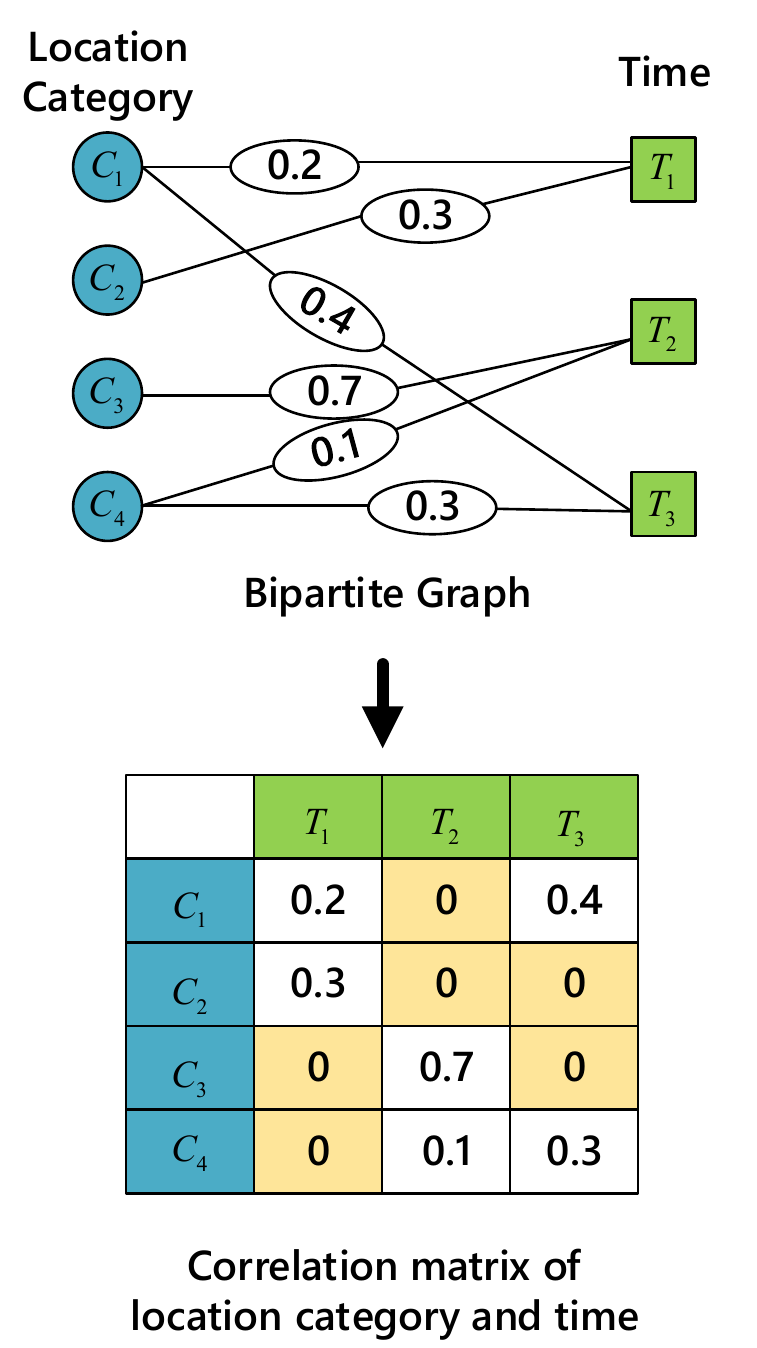}
    \caption{Using the bipartite graph to construct the activity preference matrix.}
    \label{fig.8}
\end{figure}

\textbf{Dynamic activity preference module.} Users usually tend to have different activities at different time. Fig.~\ref{fig.7}(a) and (b) tell us users more tend to work and commute on workdays, while (c) and (d) present users are more inclined to relax on weekends. Users also have different activity preferences at different time on the same day. Based on this, we propose a framework that can characterize the user's activity preference at different time. It is worth to mention that it is the first time to use graphical representation to define this problem. We attempt to construct a bipartite graph, of which location category and time are the two end nodes, while the correlation between them is the edges of the nodes at both ends. See details in Fig.~\ref{fig.8}.

We define a bipartite graph as $G=(V,E)$, where $V$ is a set containing two types of vertices: $V=V_c\cup{V_t}$, $V = \{c_1,c_2,c_3,...,c_n,t_1,t_2,t_3,...t_m\}$. $E=\{(c_i,t_j,w_{c_i,t_j})\}$, where the weight of the edge is $w_{c_i,t_j}$. $w_{c_i,t_j}$ represents the weight of $c_i$ at time $t_j$. It is a collection of edges with two different types of vertices at both ends. The task is to learn the user's activity preference at each timestamp. However, it would be expensive to directly iterate over the bipartite graph. To alleviate this problem, we generate a candidate list of all activities at each moment, and get the correlation between each timestamp and all location categories. Then we generate the weight vector between the recent state and the historical trajectory based on the bipartite graph.

\begin{equation}
    w_{c_i,t_j} = f(c\wedge{t})
\end{equation}

\begin{equation}
    \alpha{(c,t)} = \frac{exp^{w_{c,t}}}{\Sigma^{47}_{t=0}exp^{w_{c,t}}}
\end{equation}
where $w_{c,t}$ is the weight of activity $c$ at timestamp $t$ in activity preference bipartite graph. $\alpha({c,t})$ is the weight vector between recent state and the historical trajectory.

Finally, we utilize the activity preference bipartite graph to model the long and short-term activity group preference of the user,
\begin{equation}
    \bm{{\rm P}}^{c}_{L} = \bm{{\rm H}}_{h}\alpha(c,t)^{T}
\end{equation}

\begin{equation}
    \bm{{\rm P}}^{c}_{S} = \bm{{\rm H}}_{c}\alpha(c,t)^{T}
\end{equation}
where $\bm{{\rm P}}^{c}_{L}$ and $\bm{{\rm P}}^{c}_{S}$ represent long and short-term activity group preference of a user.

Using the dynamic spatio-temporal dependency module, we can capture the influence of spatio-temporal factors on users' long and short-term group preferences, which are represented  as follows respectively,

\begin{equation}
    \bm{{\rm P}}_{L} = \bm{{\rm P}}^{s}_{L} + \bm{{\rm P}}^{t}_{L} + \bm{{\rm P}}^{c}_{L}
\end{equation}

\begin{equation}
    \bm{{\rm P}}_{S} = \bm{{\rm P}}^{s}_{S} + \bm{{\rm P}}^{t}_{S} + \bm{{\rm P}}^{c}_{S}
\end{equation}

\subsection{\textbf{Multi-supervised Prediction Module}}

\begin{figure}[htpb]
    \centering
    \includegraphics[width=8cm]{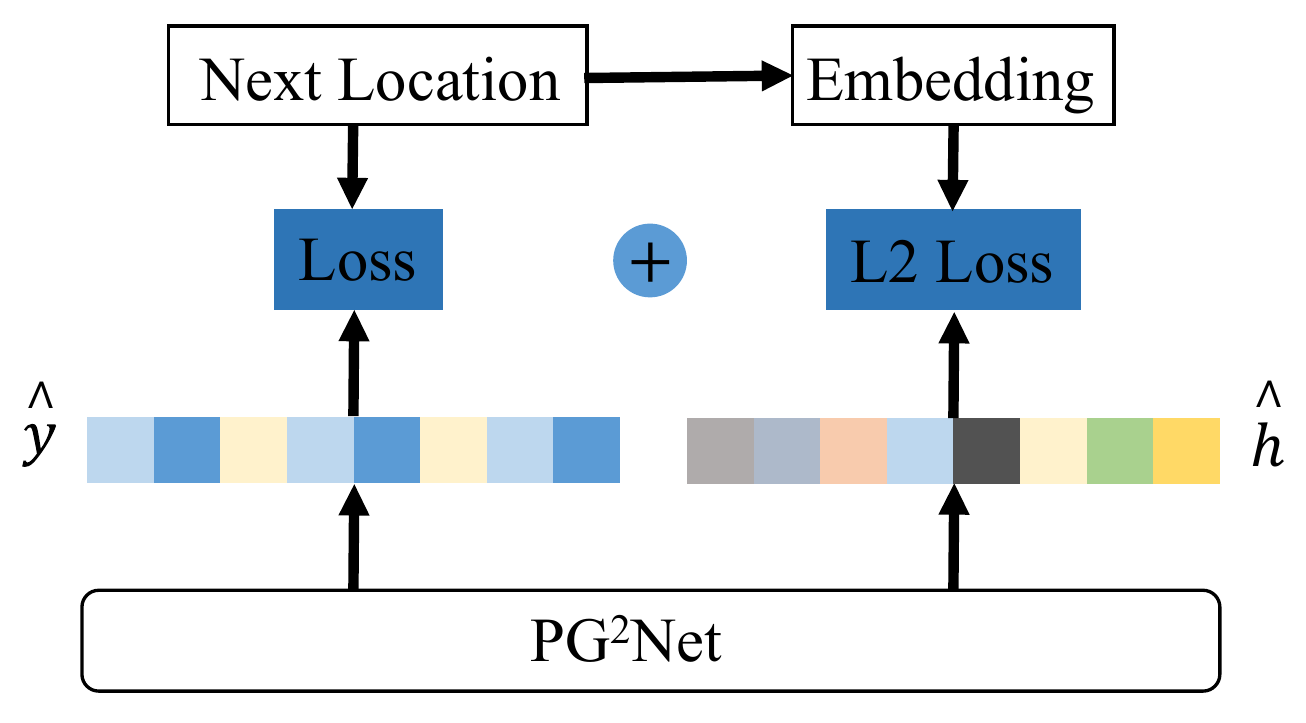}
    \caption{Multi-supervised loss function}
    \label{fig.9}
\end{figure}

After obtaining the representations for personalized preferences, long and short-term group preference, we make use of the softmax function to compute the probability distribution \bm{{\rm p}} of the next location as follows:

\begin{equation}
    \bm{{\rm p}} = softmax(\bm{{\rm W}}_p(\bm{{\rm P}}_u\oplus{\bm{{\rm P}}_L}\oplus{\bm{{\rm P}}_S}\oplus{\bm{{\rm V}}^{u}}))
\end{equation}
where $\oplus{}$ represents the concatenation of personalized preferences with long and short-term group preference, $\bm{{\rm W}}_{p}$ is a trainable matrix. Consequently, the index of the largest probability is used as the predicted value of the next location. When training the model, we use negative log likelihood as the loss function. However, in the sequence structure model, the outputted hidden state can more effectively represent the user's potential interest~\cite{zhou2019deep}. As a result, in order to improve the network's prediction accuracy, we propose an auxiliary loss function to supervise the hidden state of the user's target location. Our proposed loss function shown in  Fig.~\ref{fig.9} is defined as follows:

\begin{equation}
    L = \frac{1}{N}(-\Sigma^{N}_{i=1}log(p_i) + \varepsilon{\Sigma^{N}_{i=1}(v^{l}_{i}-\hat{h_i})^2)})
\end{equation}
where N represents the numbers of the training set, and $\varepsilon$ is a hyperparameter. We choose the L2 loss as our auxiliary loss function. $\varepsilon$ is used to balance the weight of the prediction and auxiliary loss function. With the help of the auxiliary loss function, the generated hidden vector can better express the user's interest and increase the accuracy of network prediction. 

\section{Experiments}

In this section, we proceed to evaluate the PG\textsuperscript{2}Net model on three real-world data (two check-in data and one CDRs data). We compare our proposed approach with state-of-the-art next location prediction models, and discuss the experimental results.

\subsection{\textbf{Datasets}}
We evaluate our model on the publicly available Foursquare check-in data collected from New York City (NYC) and Tokyo (TKY), which are widely used in related studies. In addition, we leverage CDRs data collected from Shanghai to evaluate our model. The check-in data contains the anonymized user ID, location id and its coordinate, location category and timestamp, while CDRs data contains the anonymized user ID, the base station ID and its coordinate, and the timestamp. Thus, we remove the embedding of category information when testing our model on CDRs data. And due to the lack of category information for CDRs, our model cannot be compared with PLSPL method. The check-in dataset collected about 10 months records in NYC and TKY via Foursquare from 12 April 2012 to 16 February 2013. Note that the temporal visitors are removed in check-in datasets via eliminating users those presented for less than two weeks. The CDRs are collected during March 2014 from 1,000 anonymized users. Table~\ref{table.1} presents the details of the three datasets. For the sparse check-in data, we first filter out the users with less than 10 records. Then, we split the trajectory of each user into multiple sub-trajectories at an interval of three days, and merge the two consecutive locations if their time interval is less than 10 mins. Next, we limit the number of sub-trajectories for each user to between 5 and 10. Sub-trajectories with less than 5 records are filtered out, and the sub-trajectories with more than 10 records are further divided into multiple trajectories. Finally, we use 80\% of each users' trajectories as the training set and the rest as testing set.

\begin{table}[htpb]
\centering
\caption{Statistics of the evaluation datasets.}
\label{table.1}
\resizebox{8cm}{!}{
\begin{tabular}{c|l|l|l}
\hline\hline
City     & \# users & \# locations & Timespan \\ \hline 
New York & 1083 & 227420   & 10 months \\ \hline
Tokyo    & 2293 & 573703   & 10 months \\ \hline
Shanghai & 1000 & 44476    & 1 months  \\ \hline
\end{tabular}
}
\end{table}

\begin{figure}[htpb]
    \centering
    \includegraphics[width=8cm]{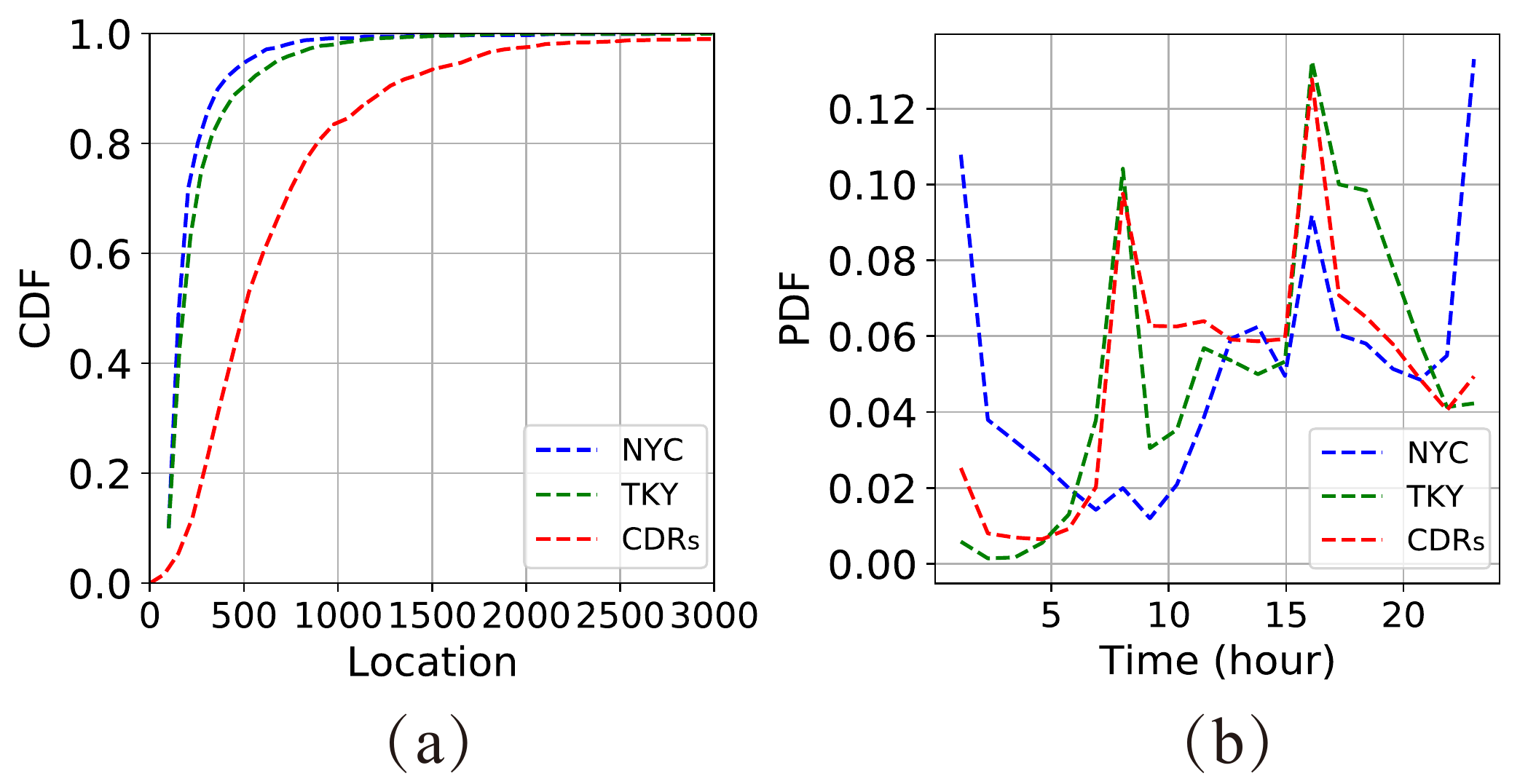}
    \caption{the statistical distribution of the number of trajectories in the check-in data and CDRs data about each user and each hour respectively.}
    \label{fig.10}
\end{figure}

More information about the three datasets is shown in Fig.~\ref{fig.10}, where (a) represents the distribution of the number of trajectories for each user, and (b) represents the proportion of the number of trajectories in each hour. We can observe from (a) that the NYC and TKY dataset are more sparse than the CDRs dataset. And (b) shows the distributions of TKY and CDRs dataset behave similarly, reflecting similar living habits of the inhabitants in Shanghai and Tokyo.

\subsection{Metrics}
For comparing our model with the baselines, we utilize two evaluation indicators: Recall@K and normalized discounted cumulative gain (NDCG@K). Recall@K measures whether there is a correct location among the top K recommended locations. NDCG@K measures the quality of top-K recommended location list. In this paper, we choose K=\{1,5,10\} for comprehensive evaluation. The definitions Recall@K and NDCG@K are given as follows,
\begin{equation}
    Recall@k = \frac{1}{N}\Sigma^{N}_{u=1}\frac{\lvert{S^{k}_u}\cap{S^{visited}_{u}}\rvert}{\lvert{S^{visited}_{u}}\rvert}
\end{equation}
\begin{equation}
    NDCG@k = \frac{1}{N}\Sigma^{N}_{u=1}\frac{1}{z_u}\Sigma^{k}_{j=1}\frac{2^{I(\{S^{j}_{u}\}\cap{S^{visited}_{u}})}-1}{log_{2}(j+1)}
\end{equation}
where $S^{k}_u$ denotes the top-k locations recommended for user u, $S^{visited}_{u}$ represents the list of locations visited in the test set, $I(\cdot)$ is an index function, $S^{j}_{u}$ represents the j-th location recommended in $S^{k}_{u}$, $z$ is the maximum value in $DCG@k$, which is a standardized constant.

\subsection{\textbf{Baselines and Settings}}
To verify the effectiveness of our proposed method, we compare PG\textsuperscript{2}Net with a classic traditional method and some mainstream deep learning methods:

\textbf{Markov Chain (MC):} It is widely used to predict human trajectories. It builds a transition matrix based on past trajectories to generate the probability of future location. In our paper, we use the first-order MC method.

\textbf{LSTM}: A neural network-based model, which is a variant model of the recurrent neural network, and can efficiently process sequence data.

\textbf{Deepmove}~\cite{feng2018deepmove}: A neural network model based on the attention mechanism that leverage each user's historical and recent trajectory data to learn her preferences. An attention mechanism is used to capture the correlation between long-term and short-term trajectory data.

\textbf{PLSPL}~\cite{wu2020personalized}: A neural network model to learn the specific preference for each user, which considers category information into the network for the first time.

\textbf{LSTPM}~\cite{sun2020go}: It is the state-of-the-art model for next location prediction, which uses context-aware non-local network structure and geo-dilated RNN to capture users’ long and short-term preferences respectively. 

\begin{figure}[htpb]
    \centering
    \includegraphics[width=7cm]{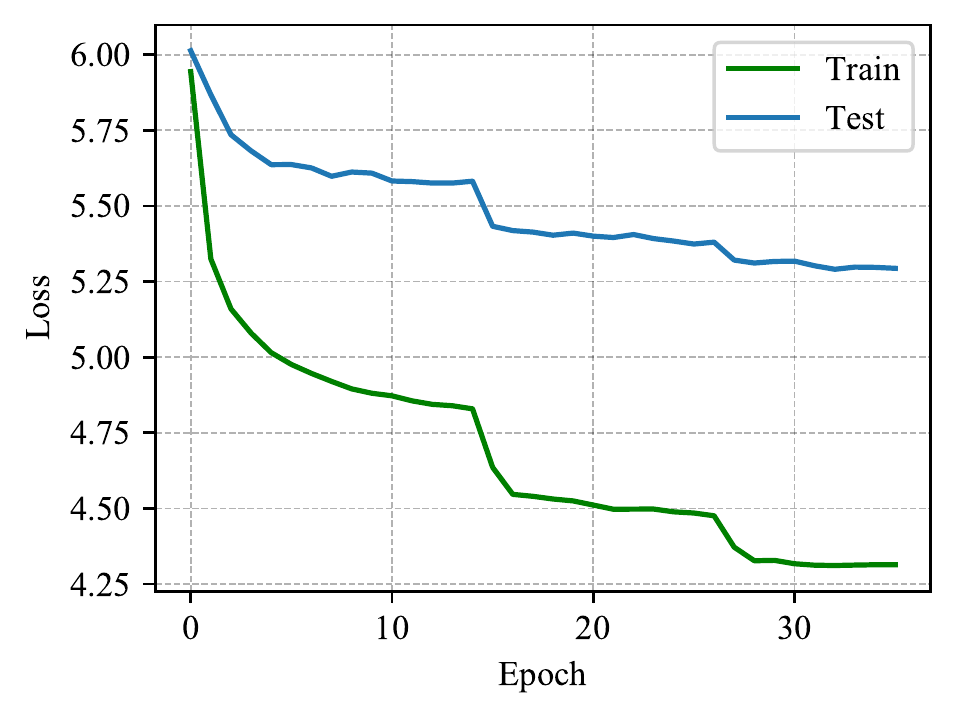}
    \caption{Loss in train and test process of TKY dataset.}
    \label{fig.11}
\end{figure}

For our method, we set the embeddings dimension of users and locations to be $D^{u} = 40$ and $D^{l} = 500$ respectively, and set categories to be $D^{c} = 50$ and $D^{t} = 10$. The dimension of the hidden state is 500. In our model we use Adam which is a gradient descent optimization algorithm to learn all the parameters. We set the initial learning rate and weight of regularization to 0.0001 and 1e-5 respectively. In the training process, we adopt the method of gradient cutting and adjust the learning rate to ensure that the model has the best performance. We take TKY dataset as an example to show the training process of the proposed model. See details in Fig.~\ref{fig.11}. For other baseline models, we set their parameters to the default values that come with the original paper.

\subsection{Result and Analysis}

\begin{table*}[htpb]
\centering
\caption{Performance comparison with five baselines on three datasets.The best method is shown in bold.}
\label{table.2}
\resizebox{\textwidth}{!}{
\begin{tabular}{lccccccc}
\multicolumn{1}{c}{Datasets} & Methods   & Rec@1  & Rec@5  & Rec@10 & NDCG@1 & NDCG@5 & NDCG@10 \\ \hline
\multicolumn{1}{l|}{}        & Markov    & 0.1356 & 0.2732 & 0.3441 & 0.1356 & 0.2078 & 0.2306  \\
\multicolumn{1}{l|}{}        & LSTM      & 0.1545 & 0.3300 & 0.3860 & 0.1545 & 0.2482 & 0.2665  \\
\multicolumn{1}{c|}{NYC}     & DeepMove~\cite{feng2018deepmove}  & 0.1828 & 0.3978 & 0.4678 & 0.1828 & 0.2967 & 0.3195  \\
\multicolumn{1}{l|}{}        & PLSPL~\cite{wu2020personalized}     & 0.1820 & 0.3947 & 0.4753 & 0.1820 & 0.2949 & 0.3210  \\
\multicolumn{1}{l|}{}        & LSTPM~\cite{sun2020go}     & 0.1864 & 0.4302 & 0.5230 & 0.1864 & 0.3143 & 0.3446  \\
\multicolumn{1}{l|}{}        & PG\textsuperscript{2}Net & \textbf{0.2120} & \textbf{0.4585} & \textbf{0.5326} & \textbf{0.2120} & \textbf{0.3437} & \textbf{0.3679}  \\ \cline{2-8} 
\multicolumn{1}{l|}{}        & Markov    & 0.1286 & 0.2500 & 0.3037 & 0.1286 & 0.1929 & 0.2102  \\
\multicolumn{1}{l|}{}        & LSTM      & 0.1440 & 0.3051 & 0.3650 & 0.1440 & 0.2293 & 0.2488  \\
\multicolumn{1}{c|}{TKY}     & DeepMove~\cite{feng2018deepmove}  & 0.1658 & 0.3609 & 0.4526 & 0.1658 & 0.2733 & 0.3034  \\
\multicolumn{1}{l|}{}        & PLSPL~\cite{wu2020personalized}     & 0.1631 & 0.3516 & 0.4294 & 0.1631 & 0.2615 & 0.2867  \\
\multicolumn{1}{l|}{}        & LSTPM~\cite{sun2020go}     & 0.1773 & 0.4052 & 0.4917 & 0.1773 & 0.2977 & 0.3258  \\
\multicolumn{1}{l|}{}        & PG\textsuperscript{2}Net & \textbf{0.1994} & \textbf{0.4336} & \textbf{0.5105} & \textbf{0.1994} & \textbf{0.3240} & \textbf{0.3490}  \\ \cline{2-8} 
\multicolumn{1}{l|}{}        & Markov    & 0.2234 & 0.4715 & 0.5520 & 0.2234 & 0.3549 & 0.3811  \\
\multicolumn{1}{l|}{}        & LSTM      & 0.2337 & 0.5516 & 0.6724 & 0.2337 & 0.3996 & 0.4390  \\
\multicolumn{1}{c|}{CDRs}     & DeepMove~\cite{feng2018deepmove}  & \textbf{0.2360} & 0.5724 & 0.6800 & \textbf{0.2360} & 0.4126 & 0.4479  \\
\multicolumn{1}{l|}{}        & LSTPM~\cite{sun2020go}     & 0.2248 & 0.5742 & 0.7016 & 0.2248 & 0.4047 & 0.4462  \\
\multicolumn{1}{l|}{}        & PG\textsuperscript{2}Net & 0.2346 & \textbf{0.5981} & \textbf{0.7021} & 0.2346 & \textbf{0.4262} & \textbf{0.4604}  \\ \hline
\end{tabular}
}
\end{table*}

The experimental results are reported in Table~\ref{table.2}. The best results in each column are highlighted in boldface. It shows that, 

(1)	The proposed model PG\textsuperscript{2}Net is compared with the baselines on three datasets, and the overall performance is superior. All the metrics on NYC and TKY datasets are better than the baseline method. On the CDRs dataset, the four measurement indicators are better than all comparison methods. Concretely, for Rec@k on NYC datasets, our method is almost 7.64\%-18.85\% higher than Markov, 5.75\%-14.66\% higher than LSTM, 2.92\%-6.48\% higher than DeepMove, 0.96\%-2.83\% higher than LSTPM. For NDGC@10, our model outperforms Markov, LSTM, DeepMove, LSTPM by 13.73\%, 10.14\%, 4.84\%, 2.33\% respectively. Our model also shows better performance than other baselines under all metrics on the TKY dataset. On the CDRs dataset, the DeepMove model performs best on Rec@1 and NDCG@1, and our model followed. Our model is the best under other metrics. The quantitative evaluation demonstrates the superior effectiveness of our method.

(2)	Among all the methods, the Markov model has the worst performance compared with other deep learning methods on the three datasets. This also shows that the neural network-based method has great advantages compared with the traditional method.

(3)	PLSPL shows better performance than LSTM on all metrics on NYC and TKY datasets. That is because PLSPL considers the context information such as category to learn the specific preference for each user. However, it shows slightly poor performance than DeepMove on NYC and TKY datasets. This phenomenon can be explained that PLSPL doesn't gather useful information from history trajectory based on the current situation.

(4)	Among the baseline methods, the LSTPM model performs best in terms of most metrics, followed by DeepMove. Compared with the DeepMove model, both our model and the LSTPM model take spatio-temporal factors into consideration, which strongly illustrates that the importance of considering the user's time and spatial factors when predicting the user's next location.

(5)	Although the deep learning method performs well on the check-in datasets, these methods don't improve the Rec@1 much when comparing with the Markov method on the CDRs dataset. We argue this probably be related to the sparsity of the trajectory data. The check-in datasets are much sparser than CDRs. For sparse data, deep learning methods can capture high-level semantic information while the traditional methods may be limited in this.

\subsection{Comparison of model variants}

\begin{table*}[htpb]
\centering
\caption{The performance comparison of next place prediction models on three datasets.}
\resizebox{\textwidth}{!}{
\label{table.3}
\begin{tabular}{lccccccc}
\multicolumn{1}{c}{Datasets} & Methods & Rec@1  & Rec@5  & Rec@10 & NDCG@1 & NDCG@5 & NDCG@10 \\ \hline
\multicolumn{1}{l|}{}        & PG\textsuperscript{2}Net   & \textbf{0.2120} & \textbf{0.4585} & \textbf{0.5326} & \textbf{0.2120} & \textbf{0.3437} & \textbf{0.3679}  \\
\multicolumn{1}{c|}{NYC}     & GNet    & 0.1795 & 0.4304 & 0.5176 & 0.1795 & 0.3116 & 0.3400  \\
\multicolumn{1}{l|}{}        & PNet    & 0.1851 & 0.4107 & 0.4932 & 0.1851 & 0.3098 & 0.3364  \\
\multicolumn{1}{l|}{}        & L-PG\textsuperscript{2}Net & 0.1918 & 0.4245 & 0.5003 & 0.1918 & 0.3159 & 0.3407  \\
\multicolumn{1}{l|}{}        & S-PG\textsuperscript{2}Net & 0.2019 & 0.4462 & 0.5227 & 0.2019 & 0.3299 & 0.3550  \\ \cline{2-8} 
\multicolumn{1}{l|}{}        & PG\textsuperscript{2}Net   & \textbf{0.1994} & \textbf{0.4336} & \textbf{0.5105} & \textbf{0.1994} & \textbf{0.3240} & \textbf{0.3490}  \\
\multicolumn{1}{c|}{TKY}     & GNet    & 0.1645 & 0.4103 & 0.4852 & 0.1645 & 0.3065 & 0.3223  \\
\multicolumn{1}{l|}{}        & PNet    & 0.1782 & 0.3951 & 0.4762 & 0.1782 & 0.3027 & 0.3203  \\
\multicolumn{1}{l|}{}        & L-PG\textsuperscript{2}Net & 0.1848 & 0.3997 & 0.4801 & 0.1848 & 0.3099 & 0.3247  \\
\multicolumn{1}{l|}{}        & S-PG\textsuperscript{2}Net & 0.1890 & 0.4224 & 0.4981 & 0.1890 & 0.3134 & 0.3379  \\ \cline{2-8} 
\multicolumn{1}{l|}{}        & PG\textsuperscript{2}Net   & \textbf{0.2346} & \textbf{0.5981} & \textbf{0.7021} & \textbf{0.2346} & \textbf{0.4262} & \textbf{0.4604}  \\
\multicolumn{1}{c|}{CDRs}     & GNet    & 0.2302 & 0.5956 & 0.6995 & 0.2302 & 0.4232 & 0.4573  \\
\multicolumn{1}{l|}{}        & PNet    & 0.2320 & 0.5926 & 0.6954 & 0.2320 & 0.4207 & 0.4545  \\
\multicolumn{1}{c|}{}        & L-PG\textsuperscript{2}Net & 0.2325 & 0.5936 & 0.6975 & 0.2325 & 0.4214 & 0.4595  \\
\multicolumn{1}{l|}{}        & S-PG\textsuperscript{2}Net & 0.2337 & 0.5970 & 0.6993 & 0.2337 & 0.4229 & 0.4565  \\ \hline
\end{tabular}
}
\end{table*}

In this section, we analyze four variants of PG\textsuperscript{2}Net to further evaluate the effectiveness of our model. The four variants are shown as follows.

\textbf{GNet}: a variant model which only engages the group preference of users, containing long-term group preference and short-term group preference.

\textbf{PNet}: a variant model which only engages the personalized preference of users, removing dynamic spatio-temporal dependency module.

\textbf{L-PG\textsuperscript{2}Net}: a variant model which engages the personalized preference and long-term group preference of users.

\textbf{S-PG\textsuperscript{2}Net}: a variant model which engages the personalized preference and short-term group preference of users.

\begin{figure}[htpb]
    \centering
    \includegraphics[width=7cm]{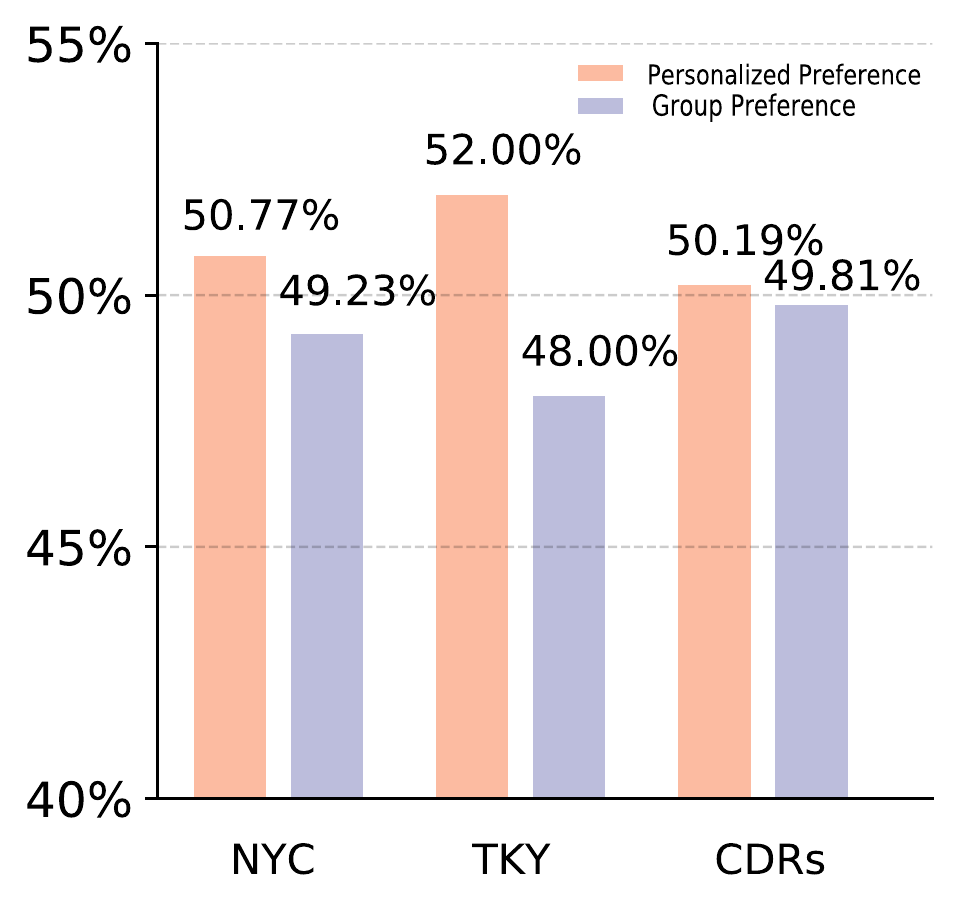}
    \caption{The weight proportion of different parts on three datasets.}
    \label{fig.12}
\end{figure}

The experimental results of the ablation tests are shown in Table~\ref{table.3}. Our PG\textsuperscript{2}Net outperforms the four variants. Specifically, (1) PNet performs better than GNet on Rec@1 and NDCG@1, while GNet performs better on other indicators. The reason is mainly on that GNet learns the group behavior pattern of all users. This pattern is generalized and can achieve a rough prediction of the next trajectory, so it performs better on Rec@5, Rec@10, NDCG@5, NDCG@10, while has a poor performance on Rec@1 and NDCG@1. For PNet, it can learn the precise and personalized preference information of each user, so the performances is better on Rec@1 and NDCG@1. This also shows that GNet and PNet can respectively achieve accurate and rough predictions of user trajectories. To further demonstrate the effectiveness of personalized preference and group preference, we compare the weight proportion of them in predicting the next location. As shown in Fig.~\ref{fig.12}, the user's personalized preference characteristics have a greater influence on the user trajectory prediction than the group characteristics. (2) S-PG\textsuperscript{2}Net always performs better than L-PG\textsuperscript{2}Net. The reason is mainly on that S-PG\textsuperscript{2}Net can better capture user’s group preferences based on her recent state. It also shows that the recent trajectory has a greater impact on the current situation.
(3) The proposed model PG\textsuperscript{2}Net, which is the combination of PNet and GNet, achieves the best performance on all test datasets. It shows that both personalized preference and group preference have positive impact on the user’s choice of the next location.

\subsection{Importance of key components in PG\textsuperscript{2}Net}

\begin{figure}[htpb]
    \centering
    \includegraphics[width=9cm]{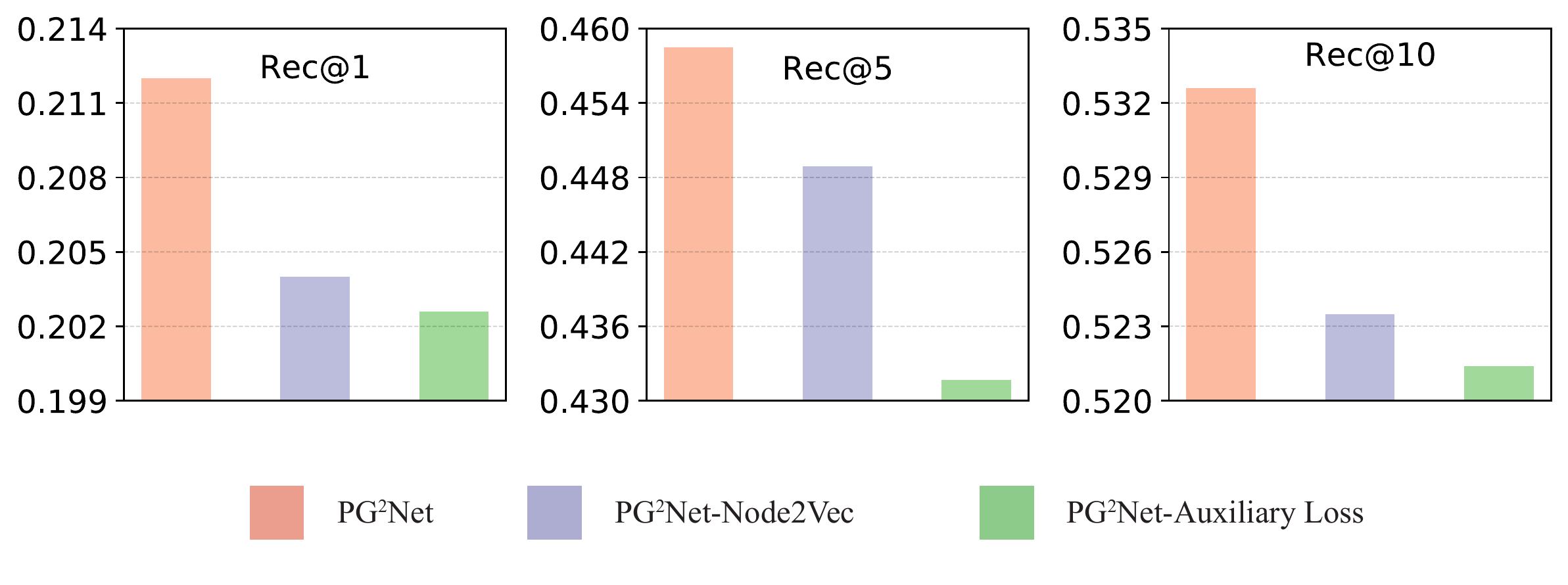}
    \caption{Analyzing on the impact of node2vec embedding and auxiliary loss}
    \label{fig.13}
\end{figure}

To better understand the influence of the node2vec embedding method and auxiliary loss function on network training, we use the NYC dataset to evaluate the performance of each module. As shown in Fig.~\ref{fig.13}, PG\textsuperscript{2}Net is our proposed complete model, PG\textsuperscript{2}Net-Node2vec denotes no graph embedding (node2vec) to embed user location and location category in the model, and  PG\textsuperscript{2}Net-Auxiliary Loss represents that the influence of the hidden state of the target location is not considered in the prediction module. Fig.~\ref{fig.13} shows that our complete model performs best, and other variant models have decreased performance. Among them, the performance of the PG\textsuperscript{2}Net-Auxiliary Loss model drops the most, with a drop of 2.68\% in Rec@5, indicating that the hidden vector of the target position has a great influence on the prediction accuracy of the next position. The second is the performance of the PG\textsuperscript{2}Net-Node2vec model, which shows that graph embedding training on location and location category can improve model performance.

\subsection{Analysis of spatial distribution of predicted locations}

\begin{figure}[htpb]
    \centering
    \includegraphics[width=9cm]{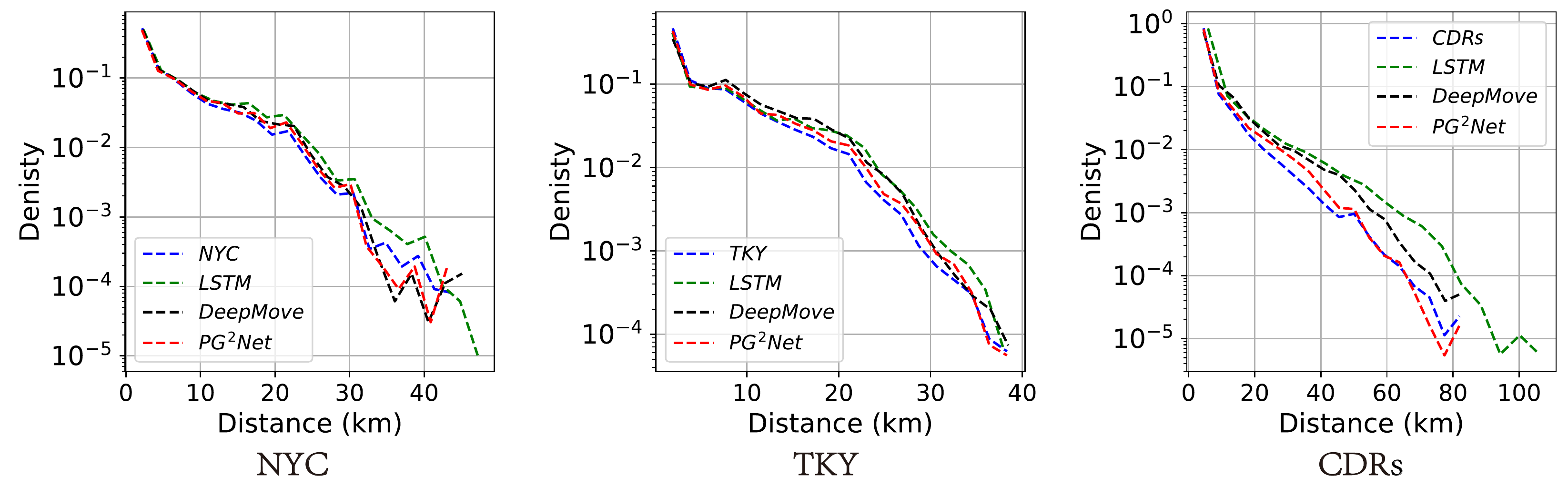}
    \caption{Distance distribution between the current situation and the next predicted location}
    \label{fig.14}
\end{figure}


To show the ability of our model to predict next locations at different distances, we examine the distance distribution between the current and the next predicted locations on the three datasets. For each dataset, we compare the actual distances with the distances predicted by the LSTM, DeepMove, and PG\textsuperscript{2}Net respectively. Fig.~\ref{fig.14} shows that when the prediction distance is short, PG\textsuperscript{2}Net has a similar prediction performance with LSTM and DeepMove. While when performing long-distance prediction, PG\textsuperscript{2}Net outperforms the LSTM and DeepMove model, and can effectively predict long-distance locations. In addition, DeepMove always outperforms LSTM on the three datasets. CDRs data can especially reflect this phenomenon. For the next places locating in less than 20~km, the performance of LSTM and DeepMove is comparable to our model. When the next places are far away, e.g., over 40~km, the performance of LSTM and DeepMove gradually deteriorates. For places locating over 80~km, the distance predicted by the LSTM model has completely deviated from the real distance distribution, and the performance is the worst, followed by DeepMove. In this scenario, the distance distribution of the predicted locations by PG\textsuperscript{2}Net matches well with the empirical data. The reasons are mainly that two locations at a long distance may be more similar for the recent state. The LSTM and DeepMove models only learn the sequence relationship of the trajectory and fails to take into account the user's personalized characteristics, the temporal and spatial information that reflects the regularities of group behavior. This could lead to LSTM and DeepMove model being unable to distinguish the two locations, resulting in long-distance jumping of the prediction locations.
\section{Conclusion and future work}
In this paper, we propose a novel end-to-end deep neural network, PG\textsuperscript{2}Net, to predict the next place to visit via considering users’ preferences to various locations at both individual and collective level. In the personalized preference module, we use Bi-LSTM and the attention mechanism to capture the users’ personalized long-term mobility tendency. In the group preference module, we use spatio-temporal and categorical information of the visited places to represent users' long-term and short-term group preferences. In addition, we utilize a graph embedding method, node2vec, to capture the sequential relation of users' visited locations and propose an auxiliary loss function to learn the vectorial representation of the target location. The extensive experimental results on three real-world datasets demonstrate the effectiveness of our proposed model. In future work, we will model more heterogeneous information and use graph neural networks to learn the interaction between them to further improve the next POI recommendation performance.

\ifCLASSOPTIONcompsoc
  \section*{Acknowledgments}
\else
  \section*{Acknowledgment}
\fi

This work was jointly supported by the Shanghai Municipal Science and Technology Major Project (2021SHZDZX0102), the Science and Technology Commission of Shanghai Municipality Project (2051102600), and the National Key Research and Development Program of China (2020YFC2008701).

\ifCLASSOPTIONcaptionsoff
  \newpage
\fi

\bibliographystyle{IEEEtran}
\bibliography{citationlist}

%







\end{document}